\documentclass{article}

\PassOptionsToPackage{numbers, compress}{natbib}

\usepackage[preprint]{neurips_2019}

\usepackage[utf8]{inputenc} %
\usepackage[T1]{fontenc}    %
\usepackage{url}            %
\usepackage{booktabs}       %
\usepackage{amsfonts}       %
\usepackage{nicefrac}       %
\usepackage{microtype}      %
\usepackage{graphicx}
\usepackage{subfigure}
\usepackage{enumitem}
\usepackage{amsmath}
\usepackage[export]{adjustbox}
\usepackage{multirow}
\usepackage{booktabs}

\newcount\Comments
\usepackage{color}
\definecolor{darkgreen}{rgb}{0,0.6,0}
\definecolor{orange}{rgb}{1,0.5,0}
\newcommand{\kibitz}[2]{\ifnum\Comments=0{\color{#1}{#2}}\fi}

\newcommand{\finalresult}{33.83}

\definecolor{citecolor}{RGB}{34, 139, 34}
\usepackage[pagebackref=true,breaklinks=true,colorlinks,bookmarks=false, citecolor=citecolor]{hyperref}

\title{Manifold Graph with Learned Prototypes for \\ Semi-Supervised Image Classification}

\author{Chia-Wen Kuo\textsuperscript{$\dagger$}, Chih-Yao Ma\textsuperscript{$\dagger$}, Jia-Bin Huang\textsuperscript{$\ddagger$}, Zsolt Kira\textsuperscript{$\dagger$}, \\
\normalsize
\textsuperscript{$\dagger$}Georgia Tech,
\textsuperscript{$\ddagger$}Virginia Tech \\
\scriptsize
\texttt{\{albert.cwkuo,cyma,zkira\}@gatech.edu},
\texttt{\{jbhuang\}@vt.edu} \\
}

\begin{document}

\maketitle

\begin{abstract}

Recent advances in semi-supervised learning methods rely on estimating the categories of unlabeled data using a model trained on the labeled data (pseudo-labeling) and using the unlabeled data for various consistency-based regularization.
In this work, we propose to explicitly leverage the structure of the data manifold based on a \textit{Manifold Graph} constructed over the image instances within the feature space.
Specifically, we propose an architecture based on graph networks that jointly optimizes feature extraction, graph connectivity, and feature propagation and aggregation to unlabeled data in an end-to-end manner.
Further, we present a novel \textit{Prototype Generator} %
for producing a diverse set of prototypes that compactly represent each category, which supports feature propagation. %
To evaluate our method, we first contribute a strong baseline that combines two consistency-based regularizers that already achieves state-of-the-art results especially with fewer labels. We then show that when combined with these regularizers, the proposed method facilitates the propagation of information from generated prototypes to image data to further improve results. We provide extensive qualitative and quantitative experimental results on semi-supervised benchmarks demonstrating the improvements arising from our design and show that our method achieves state-of-the-art performance when compared with existing methods using a single model and comparable with ensemble methods. Specifically, we achieve error rates of \textbf{3.35\%} on SVHN, \textbf{8.27\%} on CIFAR-10, and \textbf{\finalresult{}\%} on CIFAR-100.
With much fewer labels, we surpass the state of the arts by significant margins of \textbf{41\%} relative error decrease on average.
\end{abstract}

\section{Introduction}
Driven by large-scale datasets such as ImageNet and computing resources, deep neural networks have achieved strong performance on a wide variety of tasks. Training these deep neural networks, however, requires millions of labeled examples that are expensive to acquire and annotate. %
Consequently, numerous methods have been developed for semi-supervised learning (SSL), where a large number of unlabeled examples are available alongside a smaller set of labeled data. %
Most of the existing techniques for SSL fall into two categories:
1) \textit{Label transfer and pseudo-labeling}:
using predicted labels of the unlabeled data from models trained on the labeled portion~\cite{grandvalet2005semi, chen2018semi, tarvainen2017mean, Laine2017iclr, lee2013pseudo} (\textit{i.e.,} pseudo-labeling)
and label propagation~\cite{zhu2002learning, kamnitsas2018semi} based methods.
2) \textit{Consistency-based methods}: regularizing the networks using prediction consistency on unlabeled data~\cite{sajjadi2016regularization, miyato2018virtual, qiao2018deep} or ensembling model weights for better generalization~\cite{athiwaratkun2018improving}.
While some of these methods try to capture the relationships between the labeled and unlabeled data, \textit{e.g.,} through graphs, they do so where the graph structure is known~\cite{kipf2016semi, velickovic2017graph, battaglia2018relational, thekumparampil2018attention} or through multi-stage pipelines~\cite{kamnitsas2018semi}. 

\begin{figure}[t]
\vskip 0.2in
\begin{center}
    \includegraphics[width=0.75\columnwidth]{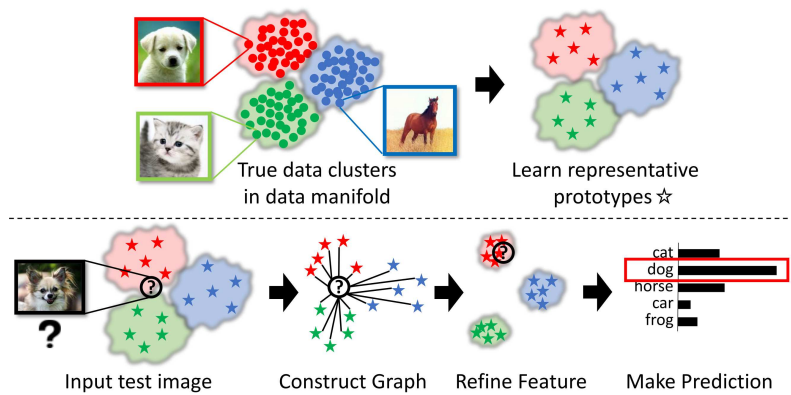}
\end{center}
\caption{
\textbf{Main idea.}
The main idea of our work is to learn a set of class-specific \textit{prototypes} during training that can compactly represent the images in the data manifold.
To classify an image, we construct a \textit{Manifold Graph} that leverages the structure of the data manifold to propagate and aggregate feature information to unlabeled data. The feature of the test image is thus refined to have a better representation that improves classification results.
}\label{fig:concept}

\vskip -0.2in
\end{figure}

In this paper, we first show that a combination of consistency-based methods (namely VAT~\cite{miyato2018virtual} and $\Pi$~\cite{sajjadi2016regularization, Laine2017iclr}) can achieve a much stronger baseline, surpassing state of the arts especially in the case where fewer labels are available.
We then propose to explicitly construct a \textbf{Manifold Graph} on the data manifold to jointly optimize feature learning, graph connectivity, and feature propagation and aggregation to unlabeled data in an end-to-end manner.
Our method builds upon a Graph Network (GN) formulation that aggregates information from neighbors to refine the feature embeddings~\cite{kipf2016semi, velickovic2017graph, kuo2019data, battaglia2018relational, thekumparampil2018attention, grover2016node2vec}.
Specifically, we represent instances (labeled or unlabeled) as nodes and use a learned similarity function on the embedding to generate the edges. %
In addition, we propose a novel \textbf{Prototype Generator} which learns to generate a set of \textit{prototypes} to compactly represent the labeled data. %
In order to learn such \textit{prototypes} effectively, we develop a set of novel loss functions that balance their effectiveness and diversity. The concept is illustrated in Fig.~\ref{fig:concept}.

Altogether, our method achieves \textbf{3.35\%} on SVHN, \textbf{8.27\%} on CIFAR-10, and \textbf{\finalresult{}\%} on CIFAR-100 (in error rate), surpassing the existing SSL approaches using a single model and comparable with ensemble methods. We further surpass all state of the arts that have been tested on more limited amounts of labeled data by significant margins, showing an average relative error rate decrease of \textbf{41.3\%}.
We will release the source code for our proposed method.
To sum up, we make the following contributions:
\begin{itemize}[topsep=0pt,itemsep=-1ex,partopsep=1ex,parsep=1ex,labelindent=0.0em,labelsep=0.2cm,leftmargin=*]
    \item We establish a strong new baseline that shows that a combination of consistency-based regularizers achieves state-of-the-art results, significantly surpassing others especially with much fewer labels.
    \item We propose a novel end-to-end graph-based method that supports feature learning as propagation and aggregation of information from learned prototypes to the image data.
    The graph connectivity is jointly optimized through a similarity function that defines an adjacency matrix tying instances that are similar on a learned embedding.
    \item We propose a novel Prototype Generator to compactly represent regions of the data manifold for each class, such that they can provide useful information within the graph network during inference in a manner that is scalable to the number of classes and prototypes.
\end{itemize}

\section{Related Work}

\paragraph{Pseudo-labeling and label propagation methods.}
Labeling-based SSL methods select \textit{confident} labels for the available unlabeled examples.
Perhaps the most intuitive approach is pseudo-labeling~\cite{lee2013pseudo}.
The prediction of unlabeled data is first computed using the model trained on the labeled data.
If the probability for a certain class exceeds a predefined threshold, it is assigned the label, otherwise discarded. 
A related idea of entropy minimization \cite{grandvalet2005semi} has also been widely adopted as an auxiliary loss combined with other SSL loss terms~\cite{miyato2018virtual}.
Other more sophisticated methods exist but require knowledge of the connectivity structure between labeled and unlabeled data, e.g., label propagation (LP)~\cite{zhu2002learning, luo2018smooth}. Unlike these method, we do not assume any knowledge of the connectivity structure and instead optimize it jointly with classification.

\paragraph{Consistency-based methods.}
Another branch of SSL methods train the network so that the prediction $\hat{y}_{u}$ of the original unlabeled data is consistent with the prediction $\hat{y}'_{u}$ of the same data after undergoing some transformation or the stochastic process of the network.
Example data perturbations include adding isotropic Gaussian random noise as in the $\Pi$ model \cite{sajjadi2016regularization} or anisotropic virtual adversarial direction as in virtual adversarial training (VAT)~\cite{miyato2018virtual}. In this work, we establish a new stronger baseline through a combination of these losses, which can already yield state of art results especially in low-label regimes. 
In addition to applying explicit data perturbation, recent work focuses on producing \emph{stable} predictions of data with unknown labels through multi-view predictions produced by multiple networks~\cite{qiao2018deep}, self-ensembling~\cite{Laine2017iclr}, or averaged model weights~\cite{tarvainen2017mean,athiwaratkun2018improving}.
In contrast, our method explicitly captures the relationship between labeled and unlabeled data samples through graph networks. Note that our method is orthogonal to the regularization losses, and therefore can adopt new advances as they occur.

\begin{figure*}[t]
\vskip 0.2in
\begin{center}
    \includegraphics[width=1\columnwidth]{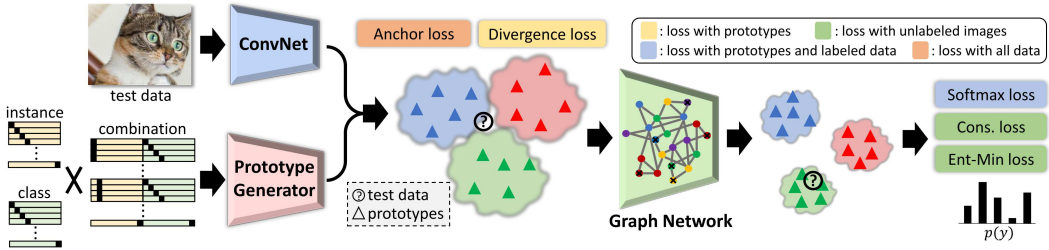}
\end{center}
\caption{
\textbf{Overview of the proposed method.}
Our proposed method leverages the structure of the data manifold, and constructs a graph to propagate information in order to refine feature representation to improve classification results.
The \textbf{Prototype Generator} learns to generate a set of prototypes, conditioned on learned instance and class embeddings, to compactly represent the data manifold.
Graph nodes are represented by an input image together with the set of generated prototypes.
The edges in the graph are learned by measuring the similarity between nodes in the embedding space.
The \textbf{Manifold Graph} module then operates on the graph to propagate feature information between nodes in the graph along the learned edges to refine the feature representations of each node.
To train this network, in addition to typical SSL losses, we introduce \textbf{anchor loss} and \textbf{divergence loss} to jointly regularize the image features and prototypes}
\label{model}
\vskip -0.2in
\end{figure*}

Lastly, the two most related approaches to our work are SSL with memory \cite{chen2018semi} and compact latent space clustering \cite{kamnitsas2018semi}.
In SSL with memory \cite{chen2018semi}, a similar idea of learning prototypes is investigated but they generate a single prototype for each class by approximating the mean of that class in the data manifold (similar to Prototypical Networks~\cite{snell2017prototypical}). A single prototype of each class may not be able to fully represent multiple modalities in the feature space.
In compact latent space clustering \cite{kamnitsas2018semi}, the authors explore the idea of building a graph to capture the underlying structure in the data manifold, similar to our work.  %
Their method however alternates between SGD and label propagation (LP) in an iterative optimization procedure per SGD iteration, which presents scalability issues.
Different from these methods, we demonstrate an end-to-end method that optimizes the entire set of components (feature extraction, graph, and prototypes) during training and allows feature propagation from the resulting prototypes to the unlabeled data. Our results demonstrates that our method can achieve better performance, while also being scalable to larger datasets such as CIFAR-100.

\section{Method}

\subsection{Baseline Method}
We select consistency-based SSL approaches as our baseline method owing to its simplicity, superior performance, and common usage in other SSL approaches \cite{Laine2017iclr, tarvainen2017mean, athiwaratkun2018improving}.
In particular, we develop a combination of the $\Pi$ model \cite{sajjadi2016regularization} and VAT perturbation \cite{miyato2018virtual}, yielding an extremely strong baseline that is already state of the art especially when there are few labeled data. The core idea of the $\Pi$ model is that given any reasonable perturbation of input image $\textbf{x}$ and stochastics of the network $f_{\theta}(\cdot)$, the new prediction distribution $\hat{y}'$ should be close to the original prediction distribution $\hat{y}$.
The consistency loss $\mathcal{L}_{cons}$ between $\hat{y}$ and $\hat{y}'$ is commonly computed using KL-divergence or mean square error.
On the other hand, VAT computes a virtual adversarial direction along which the perturbation of image $\textbf{x}$ results in the greatest change in the prediction distribution $\hat{y}$. 
We can enhance the $\Pi$ model with VAT by replacing the Gaussian noise in the $\Pi$ model with VAT perturbation.
We find this combination leads to a stronger baseline than the individual components.
In addition to consistency loss, we also introduce an entropy minimization loss $\mathcal{L}_{em}$ \cite{grandvalet2005semi} to force the network to make confident prediction as in \cite{miyato2018virtual}.
In sum, for labeled images we compute the standard cross-entropy loss $\mathcal{L}_{clf}$ for classification, and for unlabeled data we compute consistency loss $\mathcal{L}_{cons}$ and entropy minimization loss $\mathcal{L}_{em}$.

\subsection{Manifold Graph}
\label{sec:graph}
Our goal is to leverage the inherent structure in the data manifold to propagate feature representations for improving the classification of unlabeled or unseen data. 
To achieve this, we first build a fully connected graph between an unlabeled instance and training instances, or in this work their compact representations as \textit{prototypes} that we learned (see Sec \ref{sec:prototype-generator}), where the edge between each pair of nodes represents a learned similarity function between the visual features in the data manifold. 
The Manifold Graph is individually constructed by each image instance in a mini-batch of size $B$ with a set of $P$ prototypes, resulting in a graph size of $(P+1)$. 
During training, this mini-batch includes \textbf{labeled} data (to optimize features and classification through cross-entropy) and \textbf{unlabeled data} (to regularize the learning through consistency and entropy minimization losses). 
Note that we could employ a fully connected graph across the entire mini-batch and prototypes, but this would be a transductive transfer learning setting which differs from current SSL work. %

Information can then be propagated along the learned edges as a weighted sum of neighboring nodes. 
The aggregated neighboring feature representation then serves as extra information for refining the features of each node.
Our proposed Manifold Graph for information propagation is similar to a Graph Network (GN) with fixed connectivity between nodes.  
However, in our work, the lack of a predefined adjacency matrix renders this a more difficult problem, which forces us to learn both the nodes and edges jointly by optimizing with the SSL objectives.
The edges in the Manifold Graph are determined through a similarity function with a learned embedding, thereby supporting a learned adjacency matrix for the graph (similar to \cite{kamnitsas2018semi, Hsu2019iclr, ma2018attend}). 

\textbf{Learning edges:}
Similar to~\cite{vaswani2017attention}, the edge from one node can be regarded as soft attention from itself to all other nodes in the Manifold Graph.
We first map node feature $f$ to another embedding space $g = \phi(f)$ by an arbitrary embedding function  $\phi(\cdot)$.
We use a simple fully-connected layer followed by a leaky ReLU activation function here.
The edge weight between node $i$ and $j$ is then computed by taking dot product followed by softmax normalization $w_{ij} = \text{softmax}(g_i^T g_j)$.
It can be easily extended to multi-head attention by using different $\phi(\cdot)$ mapping functions for each head.
Empirically, we do not link the self-attention edge because the connectivity between one node to itself can easily dominate over other edges.

\textbf{Refining features:}
With the learned edges between each pair of nodes in the Manifold Graph, the information is then aggregated from neighboring nodes as follows:
\begin{equation}\label{msg_pass}
    h_{x_i} = \varphi \left( \left[ g_i, \sum_j w_{ij} g_j \right] \right),
\end{equation}
where $\varphi$ is a fully-connected layer and $[\cdot,\cdot]$ is concatenation operator along the feature dimension.
The 
$g_i$
carries the global information of node $i$ in the data manifold, while 
$\sum w_{ij} g_j$
encodes local contextual information around node $i$. It is crucial to include both global and local information as suggested in \cite{dgcnn}.
The original node feature is then refined by a form of residual connection $\hat{f}_{x_i} = \sigma(f_{x_i} + h_{x_i})$, where $f_{x_i}$ is the original feature of node $x_i$, and $\sigma(\cdot)$ is an activation function.

\subsection{Prototype Generator}\label{sec:prototype-generator}
\begin{figure}[t]
\vskip 0.2in
\begin{center}
    \includegraphics[width=1\linewidth]{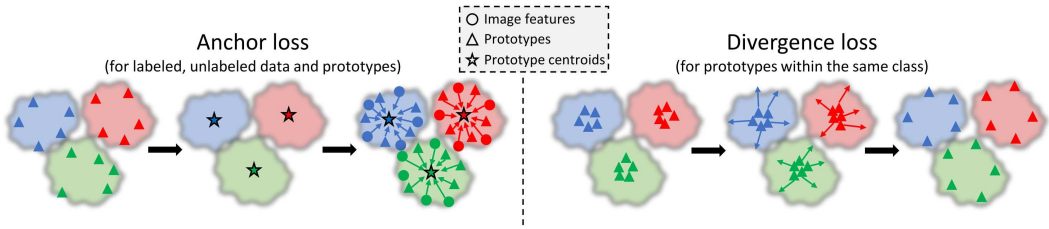}
\end{center}
\caption{
\textbf{Effect of the proposed loss functions.}
We illustrate the concept of our novel loss functions associated with the proposed Prototype Generator.
1) \textbf{Anchor loss} regularizes image features and prototypes jointly.
It aggregate these features into separable clusters in the data manifold by using the cluster centroids of each class of prototypes as anchors and decision boundaries.
2) \textbf{Divergence loss} enforces the diversity of the prototypes within each class. The goal of the divergence loss is to prevent the prototypes from collapsing.
}
\label{fig:prototype-loss}
\vskip -0.2in
\end{figure}

In Sec.~\ref{sec:graph}, we leverage the structure of the inherent manifold to propagate feature information. 
This relies heavily on the fact that the nodes in the constructed graph be sufficiently representative to capture the manifold structure.
However, the structure spanned by randomly sampled instances of labeled images may fail to capture the actual manifold structure, and is more computationally heavy as their representations must be recomputed each iteration; hence unsuitable for inference. 
To address this issue, we propose to learn compact manifold representations by generating learned \textit{prototypes}.
The learned prototypes as well as each individual image instance in a mini-batch will form a graph and be sent into our proposed Manifold Graph for joint information propagation and feature refinement. However, it is not trivial to ensure that prototypes: 1) align with the clusters of image features (to a certain controllable degree), 2) be separable across classes, and 3) be divergent enough to capture multi-modal distributions in the data manifold. 
Therefore, we propose two novel loss terms, \textit{anchor loss} and \textit{divergence loss}, to guide the learning of prototypes, as illustrated in Fig.~\ref{fig:prototype-loss}.

\textbf{Prototypes generation:}
The generator will generate $K$ prototypes per $C$ classes. We do this by generating two conditional vectors $f_k$ and $f_c$, which are simply learned embeddings that take in as input one-hot vectors of size $K$ and $C$ respectively. These one-hot vectors are then transformed to dense vector embeddings. The final conditional vector for instance $i$ of class $j$ will be:
\begin{equation}\label{proto-cond}
    f_{con, ij} = [f_{k_i}, f_{c_j}],
\end{equation}
which is simply a concatenation of $i$th embedding for $f_k$, and $j$th for $f_c$. Instead of using one single embedding with vocabulary size of $K \times C$, this greatly reduces the amount of parameters as $K$ and $C$ grow, thus scaling well to larger datasets.

Finally, the prototype $p_{ij}$ of the $i$th instance for the $j$th class can be computed as $p_{ij} =\text{MLP}(f_{con, ij})$, where  $\text{MLP}(\cdot)$ is a multi-layer perceptron. The full set of prototypes are computed for $i=[1,2,...K]$ and $j=[1,2,...C]$, thus generating $K \times C$ prototypes in total.

\textbf{Anchor loss:} In order to align the clusters of prototypes and image features, as well as separate the clusters in the data manifold based on their class, the prototypes and image features are jointly regularized by an \textit{anchor loss}.
The main idea is to first compute the cluster center for each class of prototypes, and then use these cluster centers as anchors to encourage the prototypes (and hence resulting centers) and image features to remain aligned though a triplet  and margin-based losses.

We start with computing the cluster centers $p_{c_i}$ for class $i$ as $p_{c_i} = \frac{1}{K} \sum_{k}^K p_{k,c_i}$, 
where $p_{k,c_i}$ is the $k$th prototype that belongs to class $c_i$. We use these cluster centers as anchors to regularize the generated prototypes as well as the image representations jointly. We separately regularize the \textit{magnitude}, \textit{angle}, and \textit{boundary}, which we found stabilizes the training.

\textit{Magnitude} -  This loss seeks to ensure that the prototypes do not diverge significantly in magnitude. We therefore penalize prototype centers that significantly deviate from the average feature vector length ($l_{avg}$) of images in a mini-batch. Specifically, the magnitude loss $\mathcal{L}_{mag}$ is computed as:
\begin{equation}
    \label{eq:anchor-magntitude}
    \mathcal{L}_{mag} = \frac{1}{C} \sum_{i}^{C} \text{max} \left( \left| l_{p_{c_i}}/l_{avg}-1 \right| - \text{margin}_l, 0 \right)^2,
\end{equation}
where $l_{p_{c_i}}$ is the feature length of the cluster center $p_{c_i}$, and $\text{margin}_l$ is a hyper-parameter controlling how strict these two feature lengths should match.

\textit{Angle} - In addition to magnitude, we seek to ensure that prototypes are closer (via cosine similarity) to each other (and image features) when they are of the same class than when they differ in class. We design a triplet loss~\cite{schroff2015facenet} to encourage this.
We form a triplet $(p_{c_i}, f_j, f_k)$, where $p_{c_i}$ represent the cluster center of class $c_i$, 
$f_j$ could either be an image feature or a generated prototype within the same class as $p_{c_i}$, 
and $f_k$ could also be an image feature or a generated prototype but with different class as $p_{c_i}$. Since we do not know the classes for the unlabeled data, we perform pseudo-labeling by assigning them the label of the nearest cluster centroid. 
We enforce the similarity score between the same class ($p_{c_i}$, $f_j$ pairs) to be greater than the score between different classes ($p_{c_i}$, $f_k$ pairs) by a margin. Specifically:
\begin{equation}\label{ltriplet}
    \mathcal{L}_{trip, ijk} = \text{max} (\text{S}(p_{c_i}, f_k) - \text{S}(p_{c_i}, f_j) + \text{margin}_a, 0)^2,
\end{equation}
where $\text{S}(\cdot,\cdot)$ is the cosine similarity function and $\text{margin}_a$ is a hyper-parameter controlling how separate pairs of the same class are from those of different classes.
We then enumerate all possible triplets $(i,j,k)$ and average only across non-zero terms~\cite{hermans2017defense}.

\textit{Boundary} - While the above angular loss focuses on the relative similarity between cluster centers and instances, we would also like to focus on the ``absolute'' similarity, \textit{i.e.}, decision boundary, represented by these cluster centers.
We compute the margin similarity for cluster $j$ by finding the cosine similarity of the closest cluster it:
\begin{equation}
    \text{margin}_{c_j} = \max_{i, i\neq j} \text{S}(p_{c_i}, p_{c_j}),
\end{equation}
where $\text{S}(\cdot, \cdot)$ is a cosine similarity function. We then encourage image features or prototypes $f_i$ within the class $c_j$ to be larger than this margin similarity, as follows:
\begin{equation}\label{lbound}
    \mathcal{L}_{bound, i} = \text{max}(\text{margin}_{c_j} - \text{S}(f_i, p_{c_j}), 0),
\end{equation}
where $\text{S}(\cdot, \cdot)$ is a cosine similarity function. Finally, we average over non-zero terms.
The final anchor loss $\mathcal{L}_{anc}$ is then computed as an unweighted sum $\mathcal{L}_{anc} = \mathcal{L}_{mag} + \mathcal{L}_{ang} + \mathcal{L}_{bound}$.

\textbf{Divergence loss:}
In order to avoid the prototypes from collapsing into a single mode, we also regularize the generated prototypes of the same class in a way that their similarity score is below a margin. This enforce the prototype generator to generate a diverse set of prototypes to capture the multi-modal distribution in the data manifold. %

\textit{Magnitude} - For the magnitude part, we view the average feature length $l_{avg}$ across all of the prototypes as a radius and ensure that the difference between two prototypes $(i,j)$ belonging to the same class (but where $i \neq j$) does not exceed twice this average distance:

\begin{equation}
    \label{eq:divergence-magnitude}
    \mathcal{L}_{mag, ij} = \left. \text{max} \left( \left( 1 - \frac{\mid l_i - l_j \mid}{2 l_{avg}} \right) - \text{margin}_d, 0 \right) \middle/ (1-\text{margin}_d) \right.,
\end{equation}

where $l_i$ and $l_j$ are the magnitude for prototype $i$ and $j$, and $\text{margin}_d$ is a hyper -parameter controlling the degree of divergence. Note that the maximal value in the $\text{max}(\cdot, \cdot)$ expression is $(1-\text{margin}_d)$, we thus normalize it to the range of $[0.0, 1.0]$ by dividing with $(1-\text{margin}_d)$

\textit{Angle} - Likewise, the loss of angular divergence for each pair of prototype $(i,j)$ that belongs to the same class is computed as:

\begin{equation}
    \mathcal{L}_{ang, ij} = \left. \text{max} \left( \text{S}(f_{p_i}, f_{p_j}) - \text{margin}_d, 0 \right) \middle/ (1-\text{margin}_d) \right.,
\end{equation}

where $\text{S}(\cdot,\cdot)$ is cosine similarity function. The final $\mathcal{L}_{div}$ is calculated by summing the minimum of magnitude and angular distance over all pairs $(i,j)$. %

\textbf{Summary of Loss Terms:} %
The final set of losses includes a classification loss $\mathcal{L}_{clf,I}$ for labeled images, a consistency loss $\mathcal{L}_{con}$ and entropy-minimization loss $\mathcal{L}_{em}$ for unlabeled images, and our three prototype-related loss terms ($\mathcal{L}_{anc}$, $\mathcal{L}_{div}$, and prototype classification loss $\mathcal{L}_{clf,P}$ using cross-entropy to ensure prototypes can be discriminated). 
In sum, the baseline SSL model $\mathcal{L}_{SSL}$:

\begin{equation}
\label{loss_ssl}
    \mathcal{L}_{SSL} = \overbrace{\mathcal{L}_{clf,I}}^\text{labeled} + 
    \overbrace{\lambda_1 \mathcal{L}_{con} + \lambda_2 \mathcal{L}_{em}}^\text{unlabeled}
\end{equation}

and the loss terms we have introduced to regularize the prototype generation $\mathcal{L}_{P}$:

\begin{equation}
\label{loss_proto}
    \mathcal{L}_{P} = \lambda_3 \mathcal{L}_{anc} + \lambda_4 \mathcal{L}_{div} + \lambda_5 \mathcal{L}_{clf,P}
\end{equation}

The total loss $\mathcal{L}$ is then computed as the sum $\mathcal{L} = \mathcal{L}_{SSL} + \mathcal{L}_{P}$.
We set the weights of prototype regularization ($\lambda_3$ and $\lambda_4$) to 1 and did not tune them.
We only tuned the value for the other weights on the validation set of one dataset (CIFAR-10), which we then use across all the other datasets directly.
The tuned weights are set to be $\lambda_1 = 1.0$ and $\lambda_2  = \lambda_5 = 0.1$.

\section{Experiments}

\textbf{Datasets:}
We use three standard benchmarks to demonstrate the performance of proposed method in the semi-supervised setting.
CIFAR-10~\cite{krizhevsky2009learning} is a dataset of 32x32 natural images with 50k training images and 10k test images across 10 different categories, resulting in 6k images per category. 
CIFAR-100~\cite{krizhevsky2009learning} is a similar dataset in nature except that there are 100 categories. There are also 50k training images and 10k test images like CIFAR-10, resulting in 600k images per category.
SVHN is another common dataset for SSL composed of 32x32 natural images of 0-9 digits. There are about 73k training images and 26k test images of digits for training, 26032 digits for testing. Note that we follow proper conventions described in~\cite{oliver2018realistic} for the validation set.

In the semi-supervised setting, the labeled training data is randomly separated into a smaller labeled set of $Nl$ images and a larger unlabeled set (with labels removed) of rest of the images.
We follow the common SSL setting that uses $Nl=4,000$ for CIFAR-10 , $Nl=10,000$ for CIFAR-100 , and $Nl=1,000$ for SVHN.
We also show that our model performs extremely well even with significantly less labeled data on CIFAR-10 below.

\textbf{Architecture and hyper-parameters}: Note that we do not add significant capacity, but only a few pairs of fully-connected layers (a 2\% increase in learnable parameters). All hyper-parameters were either untuned (\textit{e.g.,} prototype regularization loss weights of 1) or tuned on CIFAR-10 and used unmodified for other datasets. The exceptions are the $eps$ parameter for VAT (which was similarly tuned per dataset in the original paper) on CIFAR-100, as well as the margin$_{d}=0.9$ and margin$_{a}=0.05$ for CIFAR-100 which was chosen to account for more classes but not tuned. See Appendix for further details and discussions.

\begin{table}[t]
\centering
\small
\renewcommand{\arraystretch}{1.1}
\caption{
Quantitative comparison with state-of-the-art SSL algorithms.
We compare our implemented baseline method as well as our proposed Manifold Graph method against representative SSL methods on the SVHN, CIFAIR-10, and CIFAR-100 benchmark datasets (in error rate percentage, averaged over 3 runs with standard deviations). %
}
\label{table:sota}

\begin{tabular}{rccc}
\toprule
Method                                              & SVHN              & CIFAR-10          & CIFAR-100 \\ 
\midrule
$\Pi$-model~\cite{Laine2017iclr}                    & 4.82 $\pm$ 0.17   & 12.36 $\pm$ 0.31  & 39.19 $\pm$ 0.36 \\
NRM~\cite{ho2018neural}                             & 3.70 $\pm$ 0.04   & 11.81 $\pm$ 0.13  & 37.75 $\pm$ 0.66 \\
SSL with Memory~\cite{chen2018semi}                 & 4.21 $\pm$ 0.12   & 11.91 $\pm$ 0.22  & 34.51 $\pm$ 0.61 \\
VAT~\cite{miyato2018virtual}                        & 3.86 $\pm$ 0.11   & 10.55 $\pm$ 0.05  & - \\
Mixup~\cite{berthelot2019mixmatch}                  & 5.70 $\pm$ 0.48   & 10.26 $\pm$ 0.32  & - \\
\hline
Our VAT                                             & 3.96 $\pm$ 0.22   &  9.90 $\pm$ 0.12  & 35.06 $\pm$ 0.27 \\
Our $\Pi$-VAT                                       & 4.50 $\pm$ 0.24   &  9.20 $\pm$ 0.02  & 34.94 $\pm$ 0.24 \\
Manifold Graph (Ours)                               & \textbf{3.35} $\pm$ 0.17     & \textbf{8.27} $\pm$ 0.19      & $\textbf{\finalresult{}}$ $\pm$ 0.62 \\
\bottomrule
\end{tabular}
\end{table}

\begin{table}[t]
\centering
\small
\renewcommand{\arraystretch}{1.1}
\caption{Results with less labels on the SVHN dataset (in error rate percentage). The $\Pi$-VAT baseline that we developed is very strong. Our proposed Manifold Graph method with learned prototypes further improves this strong baseline consistently. Accuracy numbers of models we compare with are either taken from the original papers (marked with *) or from \cite{berthelot2019mixmatch}.}  
\label{table:numlabels_svhn}
\begin{tabular}{rccc}
\toprule
Method                                  & 250                       & 500                       & 1000 \\
\hline \hline
$\Pi$-model~\cite{Laine2017iclr}        & 17.65 $\pm$ 0.27          & 11.44 $\pm$ 0.39          & 8.60 $\pm$ 0.18 \\
PseudoLabel~\cite{lee2013pseudo}        & 21.16 $\pm$ 0.88          & 14.35 $\pm$ 0.37          & 10.19 $\pm$ 0.41 \\
Mixup~\cite{verma2018manifold}          & 39.97 $\pm$ 1.89          & 29.62 $\pm$ 1.54          & 16.79 $\pm$ 0.63 \\
VAT~\cite{miyato2018virtual}            & 8.41 $\pm$ 1.01           & 7.44 $\pm$ 0.79           & 5.98 $\pm$ 0.21 \\
MeanTeacher~\cite{tarvainen2017mean}    & 6.45 $\pm$ 2.43           & 3.82 $\pm$ 0.17           & 3.75 $\pm$ 0.10 \\
SSL with Memory*~\cite{chen2018semi}     & 8.83                      & 5.11                      & 4.21 \\
NRM*~\cite{ho2018neural}                 & 3.97 $\pm$ 0.21           & 3.84 $\pm$ 0.34           & 3.70 $\pm$ 0.04 \\
\hline
Our $\Pi$-VAT               & 7.26 $\pm$ 1.73           & 4.78 $\pm$ 0.41           & 4.50 $\pm$ 0.24 \\
Manifold Graph              & \textbf{3.77} $\pm$ 0.07  & \textbf{3.53} $\pm$ 0.08  & \textbf{3.35} $\pm$ 0.17  \\
\bottomrule
\end{tabular}
\end{table}
\begin{table}[t]
\centering
\small
\renewcommand{\arraystretch}{1.1}
\caption{Results with less labels on the CIFAR-10 dataset (in error rate percentage). Accuracy numbers of models we compare with are either taken from the original papers (marked with *) or from~\cite{berthelot2019mixmatch}.}
\label{table:numlabels_cifar10}
\begin{tabular}{rccccc}
\toprule
Method                                  & 250   & 500   & 1000  & 2000  & 4000 \\
\hline \hline
$\Pi$-model~\cite{Laine2017iclr}        & 53.02 $\pm$ 2.05 & 41.82 $\pm$ 1.52 & 31.53 $\pm$ 0.98 & 23.07 $\pm$ 0.66 & 17.41 $\pm$ 0.37 \\
PseudoLabel~\cite{lee2013pseudo}        & 49.98 $\pm$ 1.17 & 40.55 $\pm$ 1.70 & 30.91 $\pm$ 1.73 & 21.96 $\pm$ 0.42 & 16.21 $\pm$ 0.11 \\
MixUp~\cite{verma2018manifold}          & 47.43 $\pm$ 0.92 & 36.17 $\pm$ 1.36 & 25.72 $\pm$ 0.66 & 18.14 $\pm$ 1.06 & 13.15 $\pm$ 0.20 \\
VAT~\cite{miyato2018virtual}            & 36.03 $\pm$ 2.82 & 26.11 $\pm$ 1.52 & 18.68 $\pm$ 0.40 & 14.40 $\pm$ 0.15 & 11.05 $\pm$ 0.31 \\
MeanTeacher~\cite{tarvainen2017mean}    & 47.32 $\pm$ 4.71 & 42.01 $\pm$ 5.86 & 17.32 $\pm$ 4.00 & 12.17 $\pm$ 0.22 & 10.36 $\pm$ 0.25 \\
NRM*~\cite{ho2018neural}                 & -                & -                & 19.79 $\pm$ 0.74 & 15.11 $\pm$ 0.51 & 11.81 $\pm$ 0.13 \\
\hline
Our $\Pi$-VAT                           & 24.5 $\pm$ 2.76  & 16.76 $\pm$ 0.40 & 13.23 $\pm$ 0.10 & 11.15 $\pm$ 0.26 &  9.20 $\pm$ 0.02 \\
Manifold Graph & \textbf{19.93} $\pm$ 1.14 & \textbf{14.52}  $\pm$ 1.10 & \textbf{11.05} $\pm$ 0.52 & \textbf{9.31} $\pm$ 0.27 & \textbf{8.27} $\pm$ 0.19 \\
\bottomrule
\end{tabular}
\end{table}

\textbf{Comparison with Prior Works.} We first compare the proposed method with existing SSL approaches.
We show the results in Table~\ref{table:sota}.
As mentioned, our combination of consistency-based regularizers sets a new competitive baseline, surpassing many more sophisticated methods. Further, the quantitative results show that our proposed method consistently outperformed the existing single-model approaches on all three datasets, and even slightly surpass ensemble models on CIFAR-10 (see full table with ensemble models in Appendix).
Our approach achieves an error rate of 8.27\%, significantly outperforming the standard $\Pi$ model (12.36\% error) and VAT (10.55\% error).
Further, we achieve an absolute 3.6\% accuracy improvement on CIFAR-10 and better accuracy on CIFAR-100 over recently developed competitive state of the arts such as SSL with Memory~\cite{chen2018semi}.

Our method also achieves competitive performance when compared with methods relying on combining predictions from multiple models, either through temporal ensembling or co-training of multiple networks~\cite{Laine2017iclr, athiwaratkun2018improving, qiao2018deep}.
These methods are orthogonal to our approach and could be in principle integrated to achieve further improvement. 

\textbf{How well does our method perform with even less labeled data?} The primary goal of semi-supervised learning is to reduce the amount of labeled data while retaining comparable accuracy with the model trained with more amount of labeled data.
Therefore, we test how well our model is given even less amount of data. The results are reported in Table~\ref{table:numlabels_svhn} and Table~\ref{table:numlabels_cifar10} for SVHN and CIFAR-10 respectively.
As can be seen from the table, our $\Pi$-VAT model establishes a very strong baseline that outperforms previous best model (VAT) by a large margin. 
Nevertheless, our proposed Manifold Graph method with learned prototype still consistently improves the already strong $\Pi$-VAT model by an average of $1.5\%$.

\begin{table}[t]
\small
\renewcommand{\arraystretch}{1.1}
\centering
\caption{Ablation study on the CIFAR-10 dataset (in error rate percentage).
We verify our claim in section~\ref{sec:prototype-generator} that: (1) randomly sampled images do not serve as good prototypes, and (2) anchor loss and divergence loss are important to achieve the best classification result.}
\label{table:ablation}
\begin{tabular}{rccc|c}
\toprule
Method                  & Anchor Loss   & Divergence Loss   & Learned Prototypes    & CIFAR-10 \\
\hline \hline
random images           & -             & -                 &                       & 8.75 $\pm$ 0.17 \\
w/o anchor loss         &               & \checkmark        & \checkmark            & 8.52 $\pm$ 0.20 \\
w/o divergence loss     & \checkmark    &                   & \checkmark            & 8.32 $\pm$ 0.26\\
w/o either loss     &     &                   & \checkmark            & 8.49 $\pm$ 0.36\\
\hline
all                     & \checkmark    & \checkmark        & \checkmark            & \textbf{8.27} $\pm$ 0.19\\

\bottomrule
\end{tabular}
\end{table}

\textbf{How do the prototypes affect classification accuracy?} 
We perform an ablation study to validate our claims that learned prototypes are superior to random labeled instances and the effectiveness of our regularization, shown in Table~\ref{table:ablation}.
We can see that using randomly sampled images as prototypes significantly increases the error rate on CIFAR-10.
We also show that by combining both the anchor loss and the divergence loss, our method achieves the best accuracy. Note that the anchor loss is much more important for performance than the divergence loss, which largely achieves similar performance as without but, when combined with the anchor loss, encourages the prototypes to both be diverse and anchor to the data distribution. This can be seen in the t-SNE visualizations in Fig.~\ref{fig:qualitative} (left).

Please see Appendix for additional comparisons through qualitative and quantitative analysis, including visualization of the learned prototypes and connectivity matrix.

\section{Conclusion}
We introduce a Manifold Graph and Prototype Generator for semi-supervised image classification.
The Manifold Graph captures the structure of data manifold, allowing for jointly optimizing feature extraction, feature refinement, and label propagation in an end-to-end and scalable manner.
Rather than including labeled data for feature propagation, we instead propose the Prototype Generator, which during training learns a compact representation for each category.
The resulting prototypes support the propagation of features to the unlabeled or unseen test data.
We provide detailed quantitative evaluation that demonstrates that our prototype generation significantly improves the performance over existing methods.
Altogether, our method achieves \textbf{3.35\%} on SVHN, \textbf{8.27\%} on CIFAR-10, and \textbf{\finalresult{}\%} on CIFAR-100 (in error rate), surpassing the existing SSL approaches using a single model and comparable with ensemble methods. %
We also show that our method performs extremely well even with significantly less amount of labeled data, i.e. 1000 labeled data with only 11.55\% error rate.
Future work will combine our architecture with other SSL approaches, such as ensemble methods, which are orthogonal to our approach. %

\section*{Acknowledgement}
This work was supported by DARPA’s Lifelong Learning Machines (L2M) program, under Cooperative Agreement HR0011-18-2-0019.

\section*{Appendix}
\subsection*{Implementation Details}

\textbf{Network:} In all of the experiments, we use a standard 13-layer CNN \cite{Laine2017iclr} and drop the last fc classification layer as our feature extractor. We use leaky-ReLU with negative slope = 0.1 and dropout rate = 0.3 for the model. We use 20 prototypes per class for the Prototype Generator. We use one layer Manifold Graph module with one attention head.

\textbf{Data preparation:} For both CIFAR-10 and CIFAR-100, each mini-batch is composed of randomly sampled 32 labeled images and 128 unlabeled images, which are pre-processed by the ZCA transform. We include random horizontal flip and random translation of 2 pixels as data augmentation, which are also standard in most other SSL papers \cite{miyato2018virtual, tarvainen2017mean, Laine2017iclr}.
For SVHN, we use the same 32 labeled images and 128 unlabeled images in every mini-batch. Images are preprocessed by subtracting $mean = [0.4376821 , 0.4437697 , 0.47280442]$, and dividing by  $std =[0.19803012, 0.20101562, 0.19703614]$.
We follow the standard approach to use only random translation of 2 pixels as data augmentation.

\textbf{Training and optimization:} We use SGD optimizer paired with the learning rate (lr) scheduler of super-convergence \cite{smith2019super} for fast training. The maximal lr is chosen by the lr range test proposed in the same paper, which is 2e-2 for both datasets. We warm up training by increasing the lr from 2e-4 to 2e-3 linearly for 2,000 iterations. In the ramp-up stage, the lr increases linearly from 2e-3 to 2e-2 in 120,000 iterations, and decreases linearly to 2e-3 in another 120,000 iterations in the ramp-down stage. In the ending stage, the lr decreases from 2e-3 to 2e-5 for 40,000 iterations. We use Nesterov momentum \cite{sutskever2013importance} and set its maximal value to 0.95. On the other hand, based on the same paper, the momentum remains the same 0.95 value in the first warm-up stage as well as the last ending stage. In the ramp-up stage, the momentum linearly decreases to 0.85, and then linearly increases back to 0.95 in the ramp-down stage. This is a standard approach proposed in \cite{smith2019super} and we do not tune heavily on these parameters.

During the warm-up stage, we do not include the graph network. It pre-trains the feature extractor with a better initialization for the manifold structure, which is what the Manifold Graph relies on to compute the adjacency matrix.

\textbf{Hyperparameters:} For tuning the parameters, we construct a validation set held out from the unlabeled data in the training set. The size of validation set is 1,000 for both SVHN and CIFAR-10 as suggested by \cite{oliver2018realistic}. On the other hand, we use 2,500 for CIFAR-100 in proportion to number of labeled data we have during training.

For the VAT and $\Pi$-VAT baseline models, we use the default parameters from the original VAT paper on SVHN and CIFAR-10~\cite{miyato2018virtual}. Since the original paper does not report results on CIFAR-100, we tuned the dataset dependent parameter $eps$, which represents the length of adversarial direction. Based on the results tuned on the validation set, we set $eps=30$ for CIFAR-100. For our GN method, we also tune the $eps$ on the validation set of all three datasets, and set the value to 7, 8, and 27, for SVHN, CIFAR-10, and CIFAR-100 respectively. Finally, we increase margin$_{d}=0.9$ and margin$_{a}=0.05$ for CIFAR-100 to account for more classes.

We set our prototype regularization losses ($\lambda_3$ and $\lambda_4$) to 1 and did not tune them. We tuned the value for the other weights on the validation set of one dataset (CIFAR-10), which we also use across all datasets, as $\lambda_1 = 1.0$ and $\lambda_2  = \lambda_5 = 0.1$. The margin for the hinge losses are tuned on CIFAR-10 and transferred to SVHN and CIFAR-100. We use margin$_l=0.1$ in Eq.~\ref{eq:anchor-magntitude}, margin$_a=0.15$ in Eq.~\ref{ltriplet}, and margin$_d=0.75$ in Eq.~\ref{eq:divergence-magnitude}.

\textbf{Software and hardware configuration: } We implemented our method using the PyTorch deep learning framework, and conducted experiments on a mixture of 1080Ti, 2080Ti, and Titan Xp GPUs. 
Our source code will be released publicly.

\begin{figure}[t]
\vskip 0.2in
\begin{center}
  \begin{minipage}[b]{0.48\columnwidth}
    \includegraphics[width=1\columnwidth]{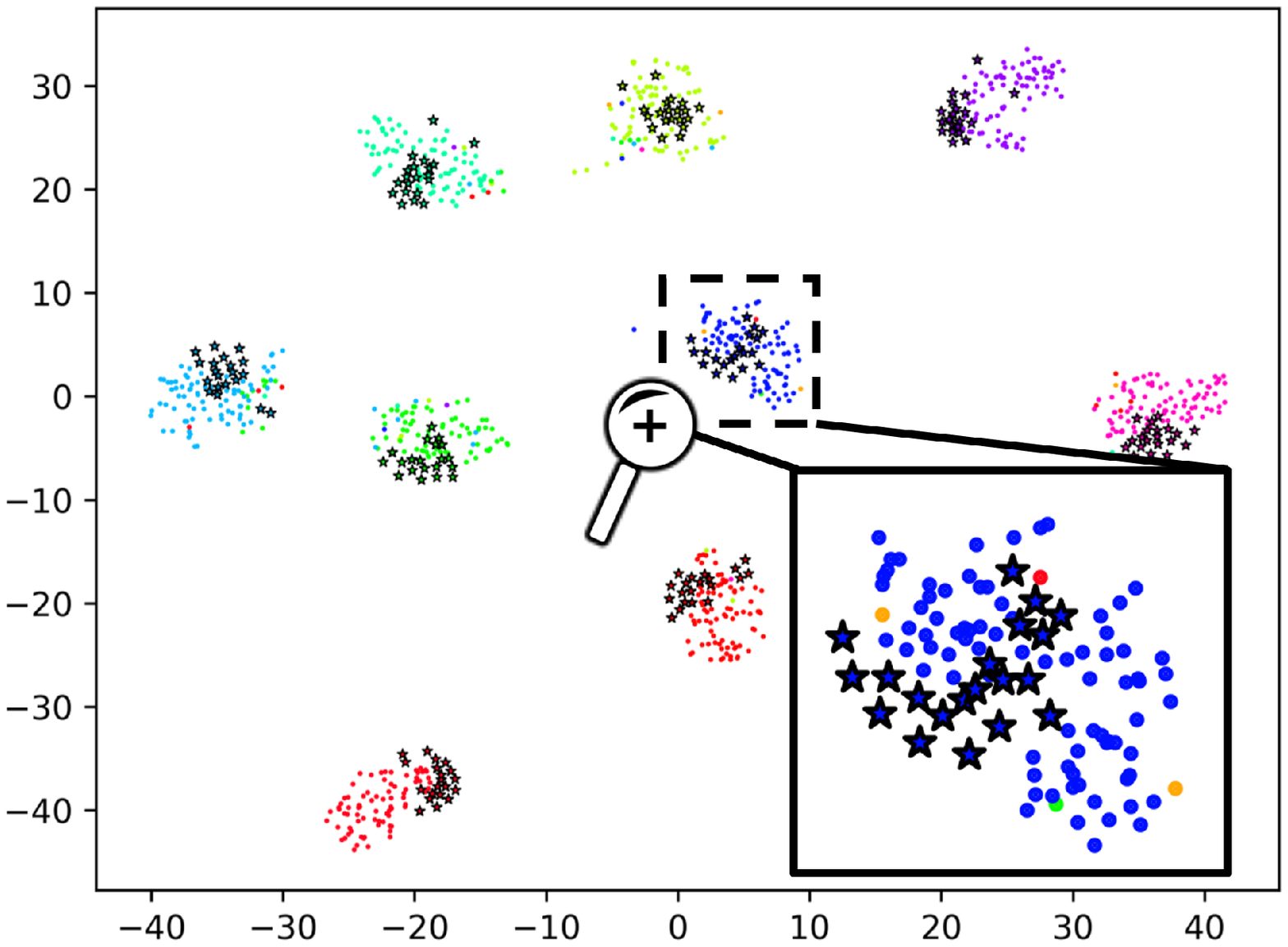}
  \end{minipage} \hfill
  \begin{minipage}[b]{0.48\columnwidth}
    \includegraphics[width=1\columnwidth]{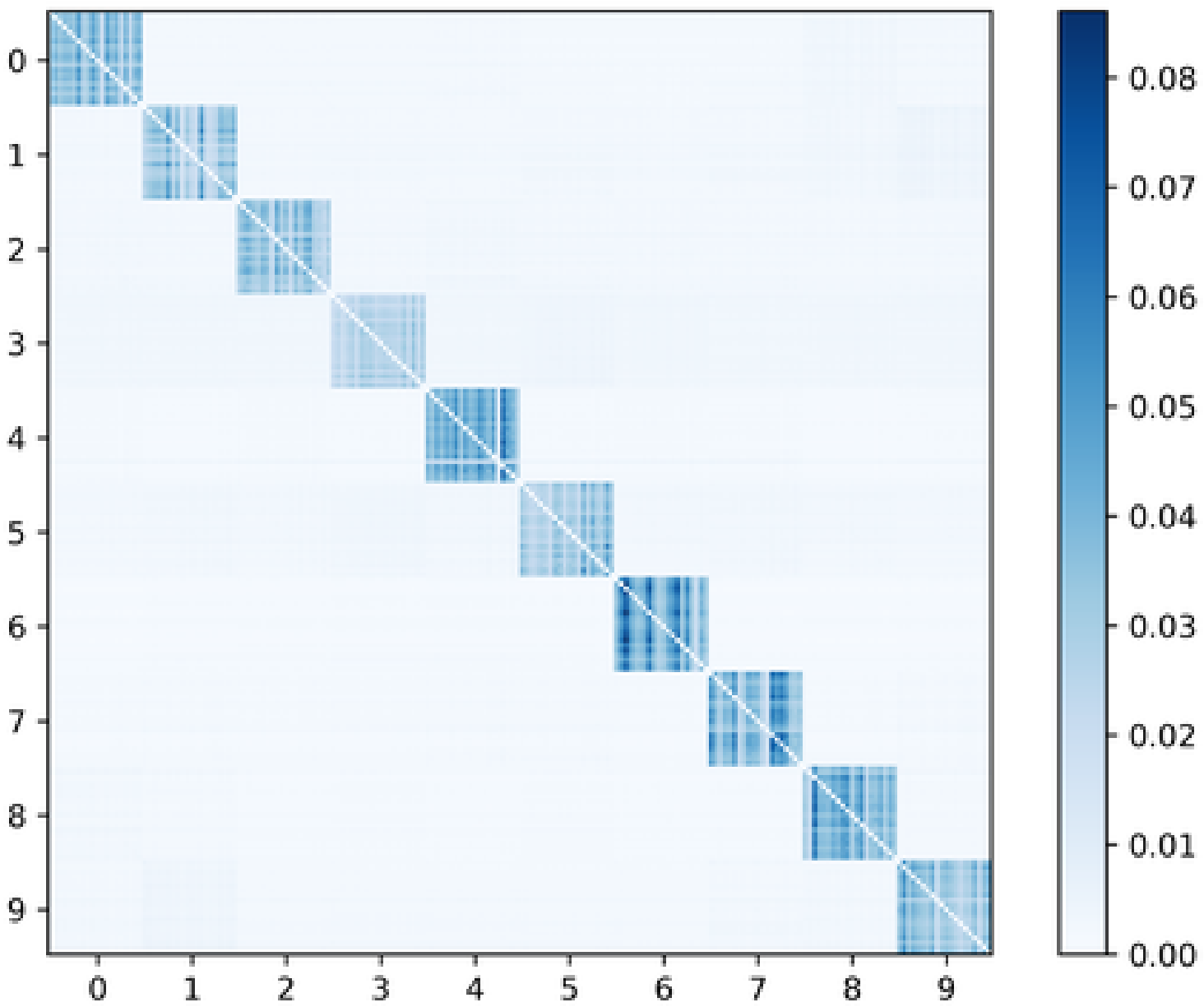}
  \end{minipage}
  \begin{minipage}{0.48\columnwidth} \centering
  Data Manifold (t-SNE)
  \end{minipage} \hfill
  \begin{minipage}{0.48\columnwidth} \centering
  Learned Connectivity (edges)
  \end{minipage}
\caption{
\textit{(left)} t-SNE visualization of the data manifold including generated prototypes as well as the entire validation set.
The stars represent prototypes, and the circles represent image features.
Same color represent the same class.
We can see that the generated prototypes align well with the clusters of image features, while retaining a good balance of diversity.
\textit{(right)} learned adjacency matrix by the Manifold Graph. Deeper colors represent stronger edge (higher edge weight).
It learns to build a stronger edge between nodes within the same class, which represents a sparser, and locally connected graph in the data manifold.}
\label{fig:qualitative}
\end{center}
\vskip -0.2in
\end{figure}

\subsection*{Additional Analysis}

\textbf{What do the prototypes look like in embedding space?}
The intuition of using prototypes is to compactly represent the data manifold for propagating and aggregating additional feature information through the Manifold Graph in order to improve the classification of the test data.
Given how the prototypes are designed with the proposed \textit{anchor} and \textit{divergence} loss, we expect the generated prototypes in the data manifold to align well with the clusters of image features for each class, while being diverse enough to capture multi-modal distributions within the same class. 
We visualize this via t-SNE (see Fig.~\ref{fig:qualitative} left).
We can see that the prototypes indeed align well with the clusters in the data manifold for test data belonging to the same class, supporting feature refinement for improved classification.
Since these prototypes carry representative feature information for each class (see Eq.~\ref{proto-cond}), the information propagated from the prototypes to the test data through the Manifold Graph improves the feature representation, hence improving the classification accuracy. One interesting finding is that our method sometimes learns prototypes \textit{along the margin}, \textit{i.e.,} between the feature distributions of different categories. This is more true when our divergence loss is not applied, since we only anchor them to the data distribution but otherwise allow them to be unconstrained. This is a potentially effective strategy, since the prototypes effectively force the graph network to project points along the margin into their correct distributions corresponding to the correct category. However, without the divergence loss the prototypes collapse and hence cannot represent a variety of prototypes.

\textbf{What edges does the Manifold Graph learn?}
The Manifold Graph is constructed on the data manifold with learned soft edges on an embedding space.
The learned edges serve as the critical path along which feature information are propagated and aggregated.
Thus, we are interested in the  graph structure that is learned by the Manifold Graph.
Note that there are no explicit loss attached to the learning of edges.
It is only jointly optimized with other loss terms. 
We thus visualize the learned adjacency matrix in Fig.~\ref{fig:qualitative} (right).
The resulting plot in the figure is colorized with deeper blue indicating higher edge weights.
The weight of the edge to itself is zeroed out as explained in Sec.~\ref{sec:graph}.
The figure shows that there is a strong block-diagonal pattern where instances within the same class tend to have much stronger edge weights while instances with different classes have significantly lower weights. 

\begin{table}[t]
\centering
\small
\renewcommand{\arraystretch}{1.1}
\caption{
Quantitative comparison with state-of-art SSL algorithms.
We compare our implemented baseline method as well as our proposed Manifold Graph method against representative SSL methods on the SVHN, CIFAIR-10, and CIFAR-100 benchmark datasets (in error rate percentage).%
}
\label{table:sota_all}

\begin{tabular}{rcccc}
\toprule
Method                                              & Ensemble  & SVHN              & CIFAR-10          & CIFAR-100 \\ 
\midrule
$\Pi$-model~\cite{Laine2017iclr}                    &           & 4.82 $\pm$ 0.17   & 12.36 $\pm$ 0.31  & 39.19 $\pm$ 0.36 \\
NRM~\cite{ho2018neural}                             &           & 3.70 $\pm$ 0.04   & 11.81 $\pm$ 0.13  & 37.75 $\pm$ 0.66 \\
SSL with Memory~\cite{chen2018semi}                 &           & 4.21 $\pm$ 0.12   & 11.91 $\pm$ 0.22  & 34.51 $\pm$ 0.61 \\
VAT~\cite{miyato2018virtual}                        &           & 3.86 $\pm$ 0.11   & 10.55 $\pm$ 0.05  & - \\
Mixup~\cite{berthelot2019mixmatch}                  &           & 5.70 $\pm$ 0.48   & 10.26 $\pm$ 0.32  & - \\
\hline
Temporal Ensembling~\cite{Laine2017iclr}            &\checkmark & 4.42 $\pm$ 0.16   & 12.16 $\pm$ 0.24  & 38.65 $\pm$ 0.51 \\
Mean Teacher~\cite{tarvainen2017mean}               &\checkmark & 3.95 $\pm$ 0.19   & 12.31 $\pm$ 0.28  & 35.96 $\pm$ 0.77 \\
SNTG~\cite{luo2018smooth}                           &\checkmark & 4.02 $\pm$ 0.20   & 12.49 $\pm$ 0.36  & 37.97 $\pm$ 0.29 \\
Weight Averaging~\cite{athiwaratkun2018improving}   &\checkmark & -                 &  9.05 $\pm$ 0.21  & \textbf{33.62} $\pm$ 0.54 \\
Deep Co-Training 2 view~\cite{qiao2018deep}         &\checkmark & 3.61 $\pm$ 0.15   &  9.03 $\pm$ 0.18  & 34.63 $\pm$ 0.14 \\
Deep Co-Training 4 view                             &\checkmark & 3.38 $\pm$ 0.05   &  8.54 $\pm$ 0.12  & - \\
Deep Co-Training 8 view                             &\checkmark & \textbf{3.29} $\pm$ 0.03   &  \textbf{8.35} $\pm$ 0.06  & - \\
\hline
our VAT                                             &           & 3.96 $\pm$ 0.22   &  9.90 $\pm$ 0.12  & 35.06 $\pm$ 0.27 \\
our $\Pi$-VAT                                       &           & 4.50 $\pm$ 0.24   &  9.20 $\pm$ 0.02  & 34.94 $\pm$ 0.24 \\
Manifold Graph (ours)                   &           & \textbf{3.35} $\pm$ 0.17     & \textbf{8.27} $\pm$ 0.19      & \textbf{\finalresult{}} $\pm$ 0.62 \\
\bottomrule
\end{tabular}
\end{table}

\subsection*{Comparison with Ensemble Methods}
We compare our method to ensembling methods including Temporal Ensembling~\cite{Laine2017iclr}, Weight Averaging~\cite{athiwaratkun2018improving}, and Deep Co-Training~\cite{qiao2018deep} in Table~\ref{table:sota_all}. We show that our proposed method outperforms Temporal Ensembling and Weight Averaging on CIFAR-10, and outperforms Temporal Ensembling and Deep Co-Training on CIFAR-100. Our method achieves comparable performance on the remaining conditions for CIFAR-100. On the other hand, our method consistently outperform those single model methods, especially on SVHN and CIFAR-10, by a large margin.

\small
\bibliography{sections/citations.bib}

\begin{thebibliography}{10}

\bibitem{athiwaratkun2018improving}
Ben Athiwaratkun, Marc Finzi, Pavel Izmailov, and Andrew~Gordon Wilson.
\newblock There are many consistent explanations of unlabeled data: Why you
  should average.
\newblock In {\em Proceedings of International Conference on Learning
  Representations}, 2019.

\bibitem{battaglia2018relational}
Peter~W Battaglia, Jessica~B Hamrick, Victor Bapst, Alvaro Sanchez-Gonzalez,
  Vinicius Zambaldi, Mateusz Malinowski, Andrea Tacchetti, David Raposo, Adam
  Santoro, Ryan Faulkner, et~al.
\newblock Relational inductive biases, deep learning, and graph networks.
\newblock {\em arXiv preprint arXiv:1806.01261}, 2018.

\bibitem{berthelot2019mixmatch}
David Berthelot, Nicholas Carlini, Ian Goodfellow, Nicolas Papernot, Avital
  Oliver, and Colin Raffel.
\newblock Mixmatch: A holistic approach to semi-supervised learning.
\newblock {\em arXiv preprint arXiv:1905.02249}, 2019.

\bibitem{chen2018semi}
Yanbei Chen, Xiatian Zhu, and Shaogang Gong.
\newblock Semi-supervised deep learning with memory.
\newblock In {\em Proceedings of the European Conference on Computer Vision
  (ECCV)}, 2018.

\bibitem{grandvalet2005semi}
Yves Grandvalet and Yoshua Bengio.
\newblock Semi-supervised learning by entropy minimization.
\newblock In {\em Advances in neural information processing systems}, 2005.

\bibitem{grover2016node2vec}
Aditya Grover and Jure Leskovec.
\newblock node2vec: Scalable feature learning for networks.
\newblock In {\em Proceedings of the 22nd ACM SIGKDD international conference
  on Knowledge discovery and data mining}, pages 855--864. ACM, 2016.

\bibitem{hermans2017defense}
Alexander Hermans, Lucas Beyer, and Bastian Leibe.
\newblock In defense of the triplet loss for person re-identification.
\newblock {\em arXiv preprint arXiv:1703.07737}, 2017.

\bibitem{ho2018neural}
Nhat Ho, Tan Nguyen, Ankit Patel, Anima Anandkumar, Michael~I Jordan, and
  Richard~G Baraniuk.
\newblock Neural rendering model: Joint generation and prediction for
  semi-supervised learning.
\newblock {\em arXiv preprint arXiv:1811.02657}, 2018.

\bibitem{Hsu2019iclr}
Yen-Chang Hsu, Zhaoyang Lv, Joel Schlosser, Phillip Odom, and Zsolt Kira.
\newblock Multi-class classification without multi-class labels.
\newblock In {\em Proc. International Conference on Learning Representations
  (ICLR)}, 2019.

\bibitem{kamnitsas2018semi}
Konstantinos Kamnitsas, Daniel~C Castro, Loic~Le Folgoc, Ian Walker, Ryutaro
  Tanno, Daniel Rueckert, Ben Glocker, Antonio Criminisi, and Aditya Nori.
\newblock Semi-supervised learning via compact latent space clustering.
\newblock In {\em International Conference on Machine Learning (ICML)}, 2018.

\bibitem{kipf2016semi}
Thomas~N Kipf and Max Welling.
\newblock Semi-supervised classification with graph convolutional networks.
\newblock {\em arXiv preprint arXiv:1609.02907}, 2016.

\bibitem{krizhevsky2009learning}
Alex Krizhevsky and Geoffrey Hinton.
\newblock Learning multiple layers of features from tiny images.
\newblock Technical report, Citeseer, 2009.

\bibitem{kuo2019data}
Chia-Wen Kuo, Jacob Ashmore, David Huggins, and Zsolt Kira.
\newblock Data-efficient graph embedding learning for pcb component detection.
\newblock In {\em 2019 IEEE Winter Conference on Applications of Computer
  Vision (WACV)}, pages 551--560. IEEE, 2019.

\bibitem{Laine2017iclr}
Samuli Laine and Timo Aila.
\newblock Temporal ensembling for semi-supervised learning.
\newblock In {\em Proc. International Conference on Learning Representations
  (ICLR)}, 2017.

\bibitem{lee2013pseudo}
Dong-Hyun Lee.
\newblock Pseudo-label: The simple and efficient semi-supervised learning
  method for deep neural networks.
\newblock In {\em Workshop on Challenges in Representation Learning, ICML},
  volume~3, page~2, 2013.

\bibitem{luo2018smooth}
Yucen Luo, Jun Zhu, Mengxi Li, Yong Ren, and Bo~Zhang.
\newblock Smooth neighbors on teacher graphs for semi-supervised learning.
\newblock In {\em Proceedings of the IEEE Conference on Computer Vision and
  Pattern Recognition}, pages 8896--8905, 2018.

\bibitem{ma2018attend}
Chih-Yao Ma, Asim Kadav, Iain Melvin, Zsolt Kira, Ghassan AlRegib, and Hans
  Peter~Graf.
\newblock Attend and interact: Higher-order object interactions for video
  understanding.
\newblock In {\em Proceedings of the IEEE Conference on Computer Vision and
  Pattern Recognition}, pages 6790--6800, 2018.

\bibitem{miyato2018virtual}
Takeru Miyato, Shin-ichi Maeda, Shin Ishii, and Masanori Koyama.
\newblock Virtual adversarial training: a regularization method for supervised
  and semi-supervised learning.
\newblock {\em IEEE transactions on pattern analysis and machine intelligence},
  2018.

\bibitem{oliver2018realistic}
Avital Oliver, Augustus Odena, Colin Raffel, Ekin~D Cubuk, and Ian~J
  Goodfellow.
\newblock Realistic evaluation of deep semi-supervised learning algorithms.
\newblock {\em arXiv preprint arXiv:1804.09170}, 2018.

\bibitem{qiao2018deep}
Siyuan Qiao, Wei Shen, Zhishuai Zhang, Bo~Wang, and Alan Yuille.
\newblock Deep co-training for semi-supervised image recognition.
\newblock In {\em Proceedings of the European Conference on Computer Vision
  (ECCV)}, 2018.

\bibitem{sajjadi2016regularization}
Mehdi Sajjadi, Mehran Javanmardi, and Tolga Tasdizen.
\newblock Regularization with stochastic transformations and perturbations for
  deep semi-supervised learning.
\newblock In {\em Advances in Neural Information Processing Systems}, 2016.

\bibitem{schroff2015facenet}
Florian Schroff, Dmitry Kalenichenko, and James Philbin.
\newblock Facenet: A unified embedding for face recognition and clustering.
\newblock In {\em Proceedings of the IEEE conference on computer vision and
  pattern recognition}, pages 815--823, 2015.

\bibitem{smith2019super}
Leslie~N Smith and Nicholay Topin.
\newblock Super-convergence: Very fast training of neural networks using large
  learning rates.
\newblock In {\em Artificial Intelligence and Machine Learning for Multi-Domain
  Operations Applications}, volume 11006, page 1100612. International Society
  for Optics and Photonics, 2019.

\bibitem{snell2017prototypical}
Jake Snell, Kevin Swersky, and Richard Zemel.
\newblock Prototypical networks for few-shot learning.
\newblock In {\em Advances in Neural Information Processing Systems}, pages
  4077--4087, 2017.

\bibitem{sutskever2013importance}
Ilya Sutskever, James Martens, George Dahl, and Geoffrey Hinton.
\newblock On the importance of initialization and momentum in deep learning.
\newblock In {\em International conference on machine learning}, pages
  1139--1147, 2013.

\bibitem{tarvainen2017mean}
Antti Tarvainen and Harri Valpola.
\newblock Mean teachers are better role models: Weight-averaged consistency
  targets improve semi-supervised deep learning results.
\newblock In {\em Advances in neural information processing systems}, 2017.

\bibitem{thekumparampil2018attention}
Kiran~K Thekumparampil, Chong Wang, Sewoong Oh, and Li-Jia Li.
\newblock Attention-based graph neural network for semi-supervised learning.
\newblock {\em arXiv preprint arXiv:1803.03735}, 2018.

\bibitem{vaswani2017attention}
Ashish Vaswani, Noam Shazeer, Niki Parmar, Jakob Uszkoreit, Llion Jones,
  Aidan~N Gomez, {\L}ukasz Kaiser, and Illia Polosukhin.
\newblock Attention is all you need.
\newblock In {\em Advances in neural information processing systems}, pages
  5998--6008, 2017.

\bibitem{velickovic2017graph}
Petar Velickovic, Guillem Cucurull, Arantxa Casanova, Adriana Romero, Pietro
  Lio, and Yoshua Bengio.
\newblock Graph attention networks.
\newblock {\em arXiv preprint arXiv:1710.10903}, 1(2), 2017.

\bibitem{verma2018manifold}
Vikas Verma, Alex Lamb, Christopher Beckham, Amir Najafi, Ioannis Mitliagkas,
  Aaron Courville, David Lopez-Paz, and Yoshua Bengio.
\newblock Manifold mixup: Better representations by interpolating hidden
  states. arxiv e-prints, art.
\newblock {\em arXiv preprint arXiv:1806.05236}, 2018.

\bibitem{dgcnn}
Yue Wang, Yongbin Sun, Ziwei Liu, Sanjay~E. Sarma, Michael~M. Bronstein, and
  Justin~M. Solomon.
\newblock Dynamic graph cnn for learning on point clouds.
\newblock {\em ACM Transactions on Graphics (TOG)}, 2019.

\bibitem{zhu2002learning}
Xiaojin Zhu and Zoubin Ghahramani.
\newblock Learning from labeled and unlabeled data with label propagation.
\newblock Technical report, Citeseer, 2002.

\end{thebibliography}
\bibliographystyle{plain}

\end{document}